# Establishing Design Routines for Efficient Control of Automated Robots

[1] Hariharan Ragothaman, [2] Harihar M, [3] S.K Guhananthan

[1, 2, 3] Department of Electrical and Electronics Engineering, Meenakshi Sundararajan Engineering College, Chennai

[1] hariharanragothaman@gmail.com
[2] m.harihar.1991@gmail.com
[3] guhasmss@gmail.com

*Abstract*— With continual development in technology, man has tried to develop robots, thereby simulating human behavior. Cognitive Robotics with AI has been efficiently used in surveying and research analysis. Though several prototypes have been created, human intervention has always been necessary, and incorporating artificial intelligence has been complicated. Through our paper, we would like to present, as to how AI can be incorporated onto robotic designs, which would ultimately improve human interactions.

Several methodologies have been proposed to enhance the robot's performance. The paper aims in designing routines for efficient control of automated robots and to test them in different environments. Digital image processing has also been introduced improving the line of sight.

One of the interesting aspects of this paper is that it tests the robotic system in real –time environments and it intends to increase their efficiencies in relative with the existing models. The paper also focuses on building a robot which has universal control and can be utilized for industrial operations as well. It has been developed and programmed on Arduino platform. The model emphasizes on Universal control to robots (GPS control), enabling safe service all times which can be utilized for surveying and research analysis. "Progressive memory algorithms" have also been simulated assisting Memory management techniques in robotics.

## I. Introduction

THIS paper primarily focuses on developing strategies for simulating the neural system onto a robot such that its performance can be enhanced. The strategies and algorithms, if implemented, would certainly bring a world of change in the field of surveying and research. The primary objective is to incorporate AI and effectively manage the fixed memory. Theoretical consciousness has also been dealt with which can pave the way for intelligent robots. Integrating progressive memory onto robots has been discussed. Self protection and awareness have been made the focal point, in addition to the design of closed loop control systems. PID controllers have been used to facilitate intelligent patterns.

The paper also discusses the modeling of a robot which has the ability to work in conditions wherein humans cannot generally dwell. The model has been designed to pick and place objects and for increasing the frequency and operating range by utilizing DTMF (Dual tone Multi Frequency) technology. This lays the foundation for survey and statistical research. Interfacing of human neural networks with robotic intelligence might seem a bit surreal but can ensure a change in the way robots are programmed and controlled. Artificial Intelligence is a mere extension of biomimicry; hence it would correspond to creating human behavioral patterns. Arduino, an open-source platform has been used to program the robot to perform the necessary steps. It is an elegant system which is user friendly, and supports various shields which helps in modeling robots for specific user defined purposes

## II. Present Scenario

### A. Progressive Memory – Intelligence

The memory of the robot is fixed and hence, as it continuously interacts with the environment there is a constant flow of data. The robot's processor is active as long it interacts with the environment. Hence the memory slowly gets exhausted as time progresses. To simulate human behavior, several modules have been installed on to the robot which occupies the already present fixed memory space. Hence the robot has limited space which it can wisely exploit for learning through experiences. As these get flooded with data feeds, overflow of data deteriorates the memory's endurance to accept further inputs. Therefore the overall efficiency of the robot reduces drastically which demands either formatting the robot's memory or replacing the robot. This would directly increase the cost involved. The humanoid's memory, in certitude, has a lot of scope for improvisation.

The brain is the ultimate repository of data; wherein information can be recalled as and when necessary. Several concepts such as the conscious and the subconscious which exist are abstract in nature. Hence the brain is considered to be active even when body is at rest. It constantly analyses data, by which it draws a relation between reality and dreams, during the period of rest. One of the most vital factors is that the brain undergoes a lot of changes, as time progresses. Since certain data stored, gets dormant in nature, the ease with which the neurons access them tends to reduce drastically. Therefore data recall after a long period of time tends to have its own share of hassles. Hence due to some form of natural degradation, no matter how much of data the human brain is fed with, there is constant amount of data loss as human's age. On comparing the processor and the brain, one of the most vital aspects to consider, is that when there is a situation of data overflow the robot rejects the excess data, while on the contrary humans accept the incoming flow of information with gradual degradation of previously existing data.

### B. Design and Interactions:-

Robotic designs involve three major factors. They are mechanical design, electronic circuitry and the programming. Fundamentally programming has entirely to do with AI which ultimately influences the other two drastically. When a robot is designed to perform a specific task, the microcontroller's programmed assuming a virtual environment which can undergo several changes under real time environments. To account for several changes, the robot has to come to a conclusion on what algorithm/task to follow when the incoming condition is not in tune with the programmed one or it does not match with the database patterns. Mechanical designs are totally driven by the surface levels and other physical parameters whereas circuitry depends on the purpose, though memory elements can be included to incorporate intelligence. Though autonomous bots exist, they face a lot of hassles on real time operations which demand human intervention. Under the current state of affairs, enhancement is mandatory, which would require better designs and AI to make automation a reality.

### C. Influence of Artificial Intelligence

AI plays a vital role in effectively performing tasks which would result in dynamic decision–making. This corresponds to discrete finalizations which could override predefined algorithms. One of the primary concerns is the driving force which would trigger the entire process. AI simplifies the entire algorithm on the whole but would look innately complex. Rudimentary modifications can make a world of difference in the operations. AI perceives actions and aims to maximize its chance of success. Sub-routines of AI are highly complex, hence interlinking of both the components appears to be surreal. If synchronized, the system would exhibit maximum stability characteristics. Arduino is an efficient tool which simplifies the entire process, thereby helping us create prototypes for the future.

### D. Human Behavioral Patterns

As mentioned earlier, the brain has an indefinite size for data storage. Sigmund Freud said "Humans tend to repress unpleasant memories". The statement vividly points out that though the brain has an astonishing ability to store huge volumes of data, paradoxically one of its primary functions is to forget. Hence the brain builds a neurobiological model of our memories with a "Guiding Enquiry" mechanism. It remembers the entire experience as one particular signal, and during the time of recall this leads to other signals, which would have previously remained dormant in nature. Using fMRI scanning process, it has been proved that an entire experience is stored as 2 simple words. We also need to analyze human behavior with respect to the environment. To begin, with actions performed can be broadly divided into explicit and implicit. It is commonly believed among humans that an act which is repeatedly performed is done better than one done implicitly. If this is the case, relative efficiency varies quite an extent; hence efficiency of an implicit operation can be boosted up only by increased activity of the brain. When a humanoid is introduced into an implicit situation, different from what it has actually programmed into, it will be pushed to a state of digital chaos. Thus the humans have succeeded through various eras' due to self-governance, which should be emulated by a humanoid/robot. The situation can be related to Darwin's theory of evolution wherein the biological entity acclimatizes itself to the variations by genetically altering itself.

### III. TURBULENCES' ON COURSE

Firstly, progressive memory has to be integrated indirectly, which has to be simulated. Further down, several mishaps may occur, but the focus is on memory constrains and self – protection. The robot might have been designed for a particular terrain, but due to some changes in the environment, it has to develop patterns of acclimatization's i.e. suitable algorithms have to be written to suit the new conditions. Turbulence of the track can pose a great threat for handling. Control design software's plays a pivotal role An image sensing interface is utilized generally. Efficient handling is mandatory for success, which can at times become futile. For example, on a rough terrain, when the robot has to move forward, Digital imaging system has to be adjusted to check the line of sight, and the motors have to be driven forward so as to move. Consider that during this entire process, a pit is encountered which is out of scope for DIM system. This would directly affect the robot and its several components.

Secondly, in the existent systems, there is no scope for progressive memory which would involve learning – working environment. Server uplinks and databases have to be initiated to match patterns. Plethoras of sensors have to be initiated to control the overall network mechanisms.

## IV. SOLUTIONS – PHASE I

### A. Simulating Learning / Working Environment

As discussed earlier during data overflow, the robot rejects data. We as humans tend to forget certain incidents due to various circumstances. If this behavior is emulated by a robot, it can have an eternal life. This sequence would involve, erasing parts of the robot's memory continually to account for more space for the constant incoming data flow. One of the obvious questions is that, how does one selectively erase one particular segment of the memory to create memory space. Since the data here, is interconnected in a lot of ways, formatting one particular area could have adverse effects on the others. The logical solution to this is, data stored can be categorically divided based on its usage. Consider a priority stack. The information frequently used can be moved towards the top section in the stack so that one can easily access, and the data generally availed to only remotely can move below, which can be shifted ahead as and when required. Since data is separated based on usage, we have defined the access flow. To deal with the problem of memory space, certain RTOS principles can be incorporated. We can have a temporary storage device, which has a sufficient amount of memory space to hold data temporarily. All the currently unused data, can move to this location, and can be retrieved to meet the requirements. By doing this, we are increasing the active memory space for the incoming data. There will be a point of saturation, wherein the size of the temporary storage area has to be increased. Hence the entire process involves in shifting data from active memory to a database and retrieving the required database as and when required via a control signal/trigger. The process by which this can possibly be simulated onto a humanoid will also be discussed in Phase II

### B. Solutions – Modules

The solutions to the discussed problems can be broadly classified into several modules. They are
1) Self-protection and awareness
2) Progressive Memory
3) Algorithm design
4) Pattern overriding
5) Human Interference Control
6) PID Theory
7) Data base Definitions

Firstly, probing onto self protection the robot experiences several obstacles /pits or other hassles which have to be overcome gradually. This can be implemented by keeping sensors and obtaining the values by which the motor/governors could be controlled. Secondly, PPA algorithms have to collaborate with the system. I.e. Perception -Processing and Action. The data flow model must be clearly analyzed so as to effectively prevent from damage. It should also be noted, process have to be comprehended in the right fashion for correct implementation.

Pattern Overriding lays the foundation of consciousness to the robot wherein the robot realizes the need to perform something over the other. This categorical discretion of one over another can implement AI onto the closed loop system. Progressive memory, if realized can be a boon to robots where they can learn while operating. Server uplinks can facilitate this process by having limited onboard memory.

Human interference Control is one of the most vital factors to be considered. This is because human- robot interactions have to be stable in order to stay away from chaos. Both the control and the controller can be stabilized through this process.

In general to integrate artificial intelligence huge databases are required to analyze the situation efficiently. Estimation and statistical analysis play a key role in this process

## V. FLOWCHARTS (CONTROLLER MECHANIZATIONS)

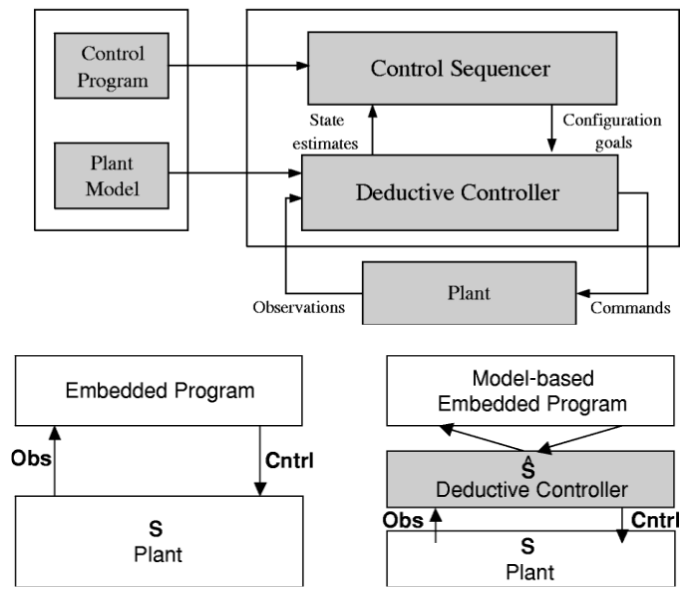

## VI. SOLUTIONS – PHASE II

### A. Utilization of Concepts:-

#### I. Primary Implementation – Progressive Memory

Let us considers a system with constant information intake. Generally the RAM is assumed to hold temporary data. Hence as the robot interacts with its surroundings, a constant feedback mechanism can be initiated to the termagant storage area. To provide an analogy, Djikstra's algorithm can be taken into consideration. It consists of weighted graph and helps in taking the minimized path. This drastically improves the flow of signals. To simplify the concept of memory management, consider a dynamic stack, wherein as data is flooded in; the top position is occupied by the data, which has just entered. With the efficient help of counters, we can suitably find the no of times the data has been accessed. Hence we can sort data in the stack. As time moves on, data will constantly be shifted to a server with expandable memory. When in need of the old data similar to frontal cortex operation of the brain, it will be on top of the stack for a given duration of time, which is obtained from the server. Later the old data is transferred to the server's safe location.

Some of the vital points in the idea conceived are, there is a constant data uplink from the humanoid's memory to the server. The server's memory to which the humanoid is connected to, will have to be expandable to serve its purpose in the long run. The server can have the capability to support a number of humanoids. Usage of a Wi-Fi Uplink maximizes the efficiency of the humanoid. Data recovery from the server will enable stack/ROM Operation. The entire process will be regenerative in nature if necessary. On the long run, the server can become full, which will demand deletion of data. To compensate for the increased cost, data from the server's memory will have to be removed to account for more space.

**Model CADD Designs – Hardware Design**

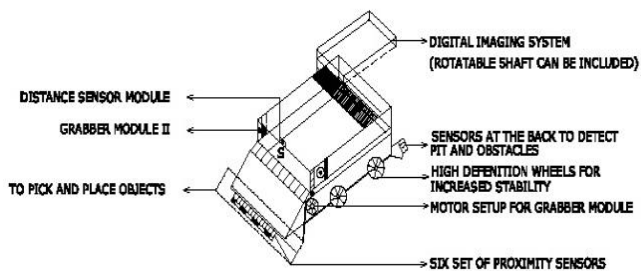

*Arduino Based Solutions:-*

**B. Arduino ~ The Open Source Software**

Arduino is an open source platform where microcontrollers can be coded using C/C++ library functions. The ATMEGA 168 is programmed through an Induino microcontroller board which has the footprint of the "Arduino" platform. The interfacing is brought via the Arduino software which effectively loads the cross platform program onto the chipset. The language used is here is primarily very much similar to JAVA which acclimatizes to the inbuilt C/C++ Library

*Case I: - Self protection*

Let's analyze each case separately to obtain solutions subsequently. To begin with self- protection should be one of the pre-requisites to be ensured. Consider the situation wherein the terrain is extremely rough which the robot has to maneuver. Proximity Sensors can be attached onto the robot to check depth and surface levels. If a pit is detected, then the robot should stop and wait for instructions which would correspond to intelligent reactive nature.

*Case II: - Human Interference Control: - A primary example:-*

Consider a robot working based on SIRC protocols. (It is controlled via a infrared remote) .Every particular address would have a particular operation to perform. In case of any danger the human's brain immediately tries to stop the ongoing action to prevent further damage. This can be efficiently exploited to suit the needs of AI .It can be programmed such that on the advent of an obstacle/pit it tends to take the reverse direction. This improves the interfacing between the two elements.

*Case III: - Progressive Memory Simulation*

To perform this we need to link the robot to the server either via an Ethernet shield / through ZIGBEE control network which could constantly access the database and receive commands so as to perform efficiently. The RAM available can be used to perform routine, repetitive tasks.

*Case IV: - Algorithm Design & PID theory*

Algorithms can be either based on rote logic /fuzzy logic. Probabilistic theories can also be conceived to improve the efficiency of the system. PID (Proportional Integral and Differential Controller) can be put to use to control the speed –torque ratios of the machine. Sharp turns can easily be recognized; i.e. the motor turns slowly during sharp ones and increases its speed as the curvature reduces. This acts as a safety mechanism. This can be done by having many proximity sensors and reading their O/P's at specific time intervals and suitably adjusts motor directions, to turn.

The same concept can be applied onto servo motors which are utilized to pick objects or collect samples. This can be done by controlling the governor of the shaft for the operation, by checking how far the object actually is with the help of ping sensors.

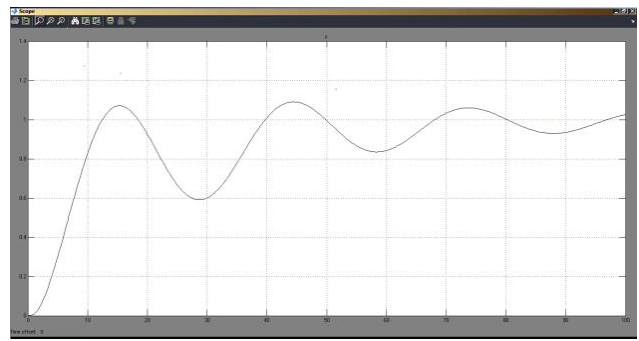

### VII. ROBOTIC DESIGN - HARDWARE

The overall setup consists of a chassis with a castor wheel, coupled with DC motors for movement. The DC motors function at 12 Volts and deliver 160 rpm at no load condition.

For the purpose of arm control DC motors have been used with a rating of 60 rpm. This is done, just to emulate the operation, which can be improvised by the use of servo's or stepper motors. The overall electronic/microcontroller board setup has been placed to provide enough room for arm movement. The ATMEGA 168 is programmed through an Induino microcontroller board which has the footprint of the "Arduino" platform. The interfacing is brought via the Arduino software which effectively loads the cross platform program onto the chipset.

The DTMF controlled grabber robot can be split into a variety of four different modules viz.
 a. The grabber
 b. The DTMF circuitry programmable interfacing
 c. SIRC Control
 d. Optical Mouse

Using the Arduino software DC motors have been controlled to operate on gears which operate in a synchronous fashion. They are placed in such a way that one motor rotates in clockwise while the  Secondly the analog O/P's of the DTMF are given as I/P's to the board. The addresses received are suitably governed to activate and manipulate the DC motors for navigational purposes. SIRC protocol is adopted to control the grabber mechanism via TSOP receivers. They operate at well defined frequency limits and can be used for short range applications .Programming the three modules and assimilating them together has been done by using 2 separate Arduino boards. The boards function independently from each other. The interfacing is done via a hassle free process enabling faster transmission and distribution.

An optical mouse has been exploited to serve our needs. We have used its inbuilt camera to track it's coordinates at any instant of time. This helps in tracking and remote sensing the robot. This is done by placing the optical mouse beneath the bot such that it faces the ground. The output of the mouse is given as input to the boards, whose co-ordinates are either printed onto an LCD screen or is directed to the server.

The code is quite efficient due to its reusability and the incorporation of object oriented nature. Once uploaded the code gets stored onto the onboard RAM. The overall efficiency of the interfacing is due to the Arduino-platform and the simplified code/syntax needed.

### VIII. PSEUDO CODE FOR TRACK ANALYSIS
### (Robo- Mind Code)

PROGRAM :-
```
repeat()
{
if(not leftIsObstacle()
and leftIsWhite())
{ right()
 backward(1)
end }
else
{ forward(1)
}
```

### IX. ELECTRONIC CIRCUITRY – SOLUTIONS – UNIVERSAL CONTROL

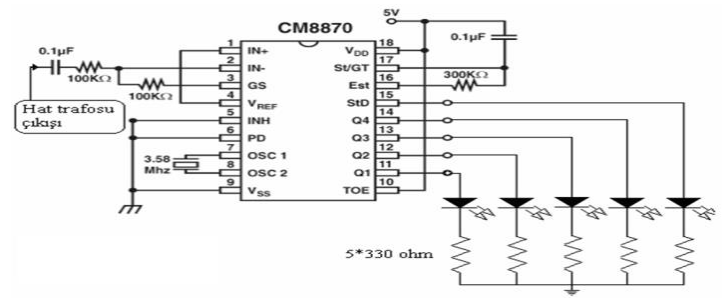

Dual-tone multi frequency is used for telecommunication signaling over analog telephone lines in the voice-frequency band telephone handsets and other communications devices and the switching center. The DTMF keypad is laid out in a 4x4 matrix, with each row representing a low frequency and each column representing a high frequency. The multiple tones generated give its name, which is read by a switching center for control operation.  This property enables it to be controlled from any part of the world.

Two phones can be linked via this technology thereby extending maximum bandwidth and increased performance levels. One of the phones is attached to the board which constantly receives data from a remote location. Hence this phone can be accessed as long as the service provider's network is active. In remote applications one can set up their own localized networks to suit their needs. This is how the universal control algorithm has been simulated.

## X. SOFTWARE SIMULATIONS

Probing into the code for implementing the software, we have utilized 2 boards and one motor shield to project our ideas. This is just a prototype, on which many modifications can be made. Firstly, as we click a particular button on the phone, it has a specific frequency, which is given as input to the board. This frequency is analyzed and converted into analog/ binary values for performing tasks.

Based on the analog values obtained, we have setup suitable threshold levels, so that a task is performed as and when the condition is satisfied. The task is just to mobilize the robot in all directions. Secondly, for the purpose of sensing obstructions, we have installed a number of proximity sensors and ping sensors which constantly read data in the vicinity thereby sending in I/Ps to the board. These are analyzed and suitable algorithms have been implemented.

As stated earlier, to serve the needs of the industry, we have designed arms for the robots which work via DC motors of 60rpm. These synchronously work to grab/ dispose an object.

**Salient Features.** :-
1. It has universal control via DTMF technology
2. Sensors attached help in simulating elementary concepts of AI
3. ZIGBEE /Ethernet shields can be utilized to link with servers to ensure overall control.
4. Optical Mouse Hack enables tracking/ sensing of the robot.
5. Elegant design and Simplified procedures.

## XI. SIMULATIONS & FLOWCHARTS

SEARCH AND TAG MODULES (ROBO-MIND SOFTWARE)

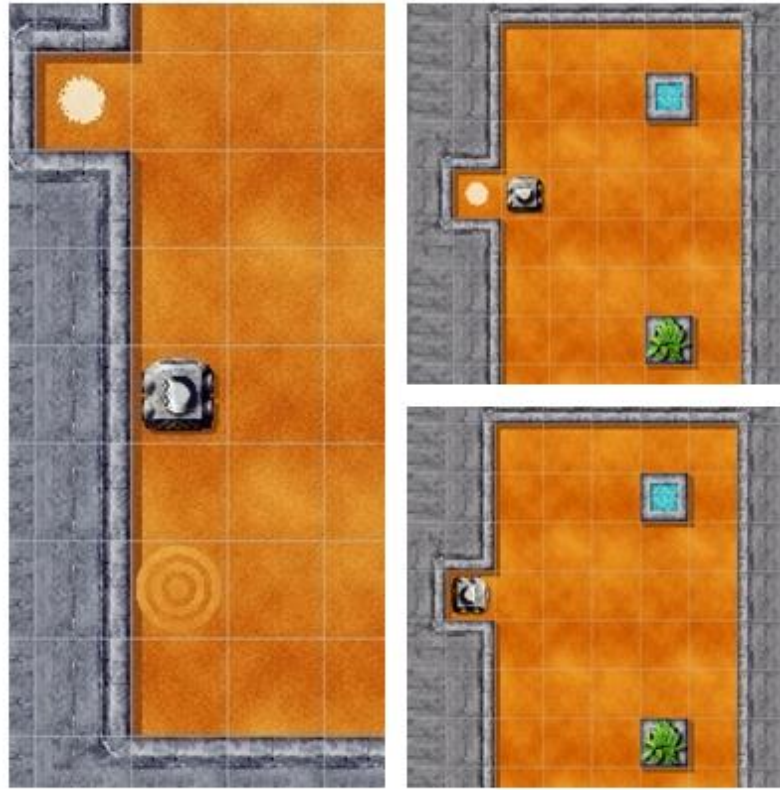

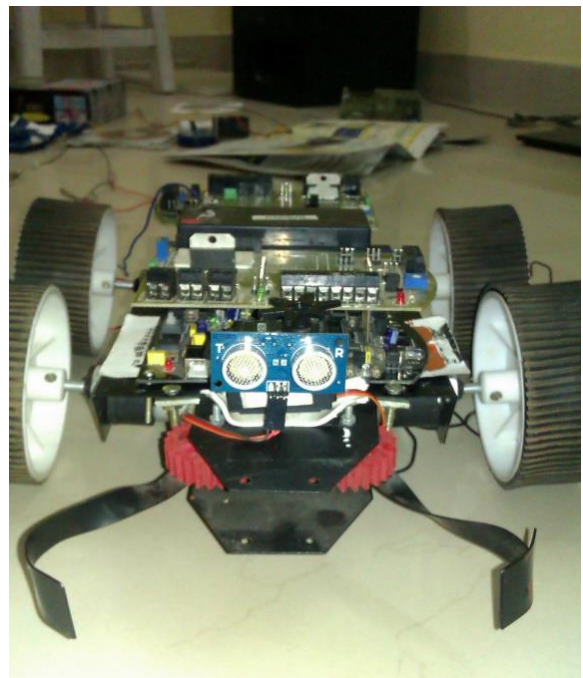

## XII. ALTERNATIVE SOLUTIONS AND STRATEGIES

In general, robots are built for a particular purpose, and can serve a larger need. Hence custom made robots' entire interaction will remain within a particular field of operation. Hence instead of linking it to a server, memory modules can be efficiently used so as to reduce the complexity involved in the operation. Thus purpose specific bots with installed chips can be used in Special Forces and medical research.

With exponential rise of robots, there will be a stage when personal robots will become come to the forefront. Currently only a part of the technical society is able to comprehend the need for personal robots.

## XIII. SCOPE FOR IMPROVEMENT & APPLICATIONS

Even though the robot is programmed for a particular task it is indeed the abstractness which stimulates imagination. A robot designed for a particular task with predefined instructions can improvise itself due to some random protocols which can be realized by unexplainable reasons. Hence it is indeed possible to stimulate the ghost within, by introducing it to random explicit situations.

Another important factor to be remembered is that at no point is the server to be introduced to the concept of Artificial Intelligence as in reality, since the server possesses the ultimate control, overruling decisions can be administered which may result in total destruction of the entire network.

The server should just be a means of storage and transfer of data. The robot can be linked to servers to add server based operations to be available for a wider range of network. Artificial Intelligence, which is one of the rising fields along with robotics, can be simulated via these systems efficiently. The controller used can be upgraded in accordance to the latest trends for wider range of address access and easier control. The overall robotic design can be modified to suit purpose specific applications.

The robot has a wide range of uses as in rescue and search operations. Remote sensing applications can be enhanced if our ideas are incorporated onto their protocols. Land rover projects, generally acid tested in space stations can utilize our concepts/features. On the account of a disaster, to calculate statistics, sensors can be embedded onto the proto-boards to facilitate purpose specific operations. The DTMF technology, so used can be improvised or rather modified as Speech convertible mediums to simulate closed loop systems.

## XIV. CONCLUSION

This paper's vision is to enable robots to operate on terrains which have been determined as hostile and to collect data/samples. A plethora of other ideas can be generated, by keeping our ideas as the base in the future. I do hope that future has a lot in store, and one day man will develop capabilities to incorporate several of the features proposed, and recreate human behavioral patterns.

## ACKNOWLEDGMENT

We would like to thank S.B Sivasubramaniyam (Asst Professor), (EEE department) for his constant support and encouragement. His constant encouragement has been the driving forces behind this paper. We also thank our department for providing us with this opportunity.